\documentclass[10pt,twocolumn,letterpaper]{article}

\usepackage{cvpr}              %

\usepackage{times}
\usepackage{epsfig}
\usepackage{graphicx}
\usepackage{amsmath}
\usepackage{amssymb}

\usepackage[accsupp]{axessibility}
\usepackage{graphicx}
\usepackage{amsmath}
\usepackage{amssymb}
\usepackage{booktabs}
\usepackage{lipsum} 
\usepackage{multirow}

\usepackage{tikz}
\usepackage{comment}
\usepackage{dsfont}
\usepackage{color, colortbl}
\usepackage{threeparttable}
\usepackage{tabularx}
\usepackage{caption}
\usepackage{subcaption}
\usepackage{hhline}
\usepackage{sidecap}
\usepackage{algorithm}
\usepackage{algorithmic} %
\usepackage{listings}
\usepackage{xcolor}

\usepackage{footnote}
\makesavenoteenv{algorithm}

\usepackage[pagebackref,breaklinks,colorlinks]{hyperref}

\definecolor{shapecolor}{RGB}{1, 128, 0}
\usepackage{etoolbox}
\makeatletter
\AfterEndEnvironment{algorithm}{\let\@algcomment\relax}
\AtEndEnvironment{algorithm}{\kern2pt\hrule\relax\vskip3pt\@algcomment}
\let\@algcomment\relax
\newcommand\algcomment[1]{\def\@algcomment{\footnotesize#1}}
\renewcommand\fs@ruled{\def\@fs@cfont{\bfseries}\let\@fs@capt\floatc@ruled
  \def\@fs@pre{\hrule height.8pt depth0pt \kern2pt}%
  \def\@fs@post{}%
  \def\@fs@mid{\kern2pt\hrule\kern2pt}%
  \let\@fs@iftopcapt\iftrue}
\makeatother

\usepackage[capitalize]{cleveref}
\crefname{section}{Sec.}{Secs.}
\Crefname{section}{Section}{Sections}
\Crefname{table}{Table}{Tables}
\crefname{table}{Tab.}{Tabs.}

\newcolumntype{C}{>{\centering\arraybackslash}X}

\def\confName{CVPR}
\def\confYear{2024}

\begin{document}

\newcommand{\ourmethod}{Plug and Play Active Learning}
\newcommand{\ourmethodFull}{Plug and Play Active Learning}
\newcommand{\ourmethodIntro}{Plug and Play Active Learning (PPAL)}
\newcommand{\ourmethodShort}{PPAL}

\newcommand{\ourUncetaintySample}{Difficulty Calibrated Uncertainty Sampling}
\newcommand{\ourUncetaintySampleFull}{Difficulty Calibrated Uncertainty Sampling (DCUS)}
\newcommand{\ourUncetaintySampleShort}{DCUS}

\newcommand{\ourSim}{Category Conditioned Matching Similarity}
\newcommand{\ourSimFull}{Category Conditioned Matching Similarity (CCMS)}
\newcommand{\ourSimShort}{CCMS}

\title{Plug and Play Active Learning for Object Detection}

\author{Chenhongyi Yang$^{1}$\thanks{Corresponding Author. Email: chenhongyi.yang@ed.ac.uk} \quad\quad  Lichao Huang$^{2}$ \quad\quad Elliot J. Crowley$^{1}$ \\
$^{1}$School of Engineering, University of Edinburgh \\
$^{2}$Horizon Robotics\\}

\maketitle

\begin{abstract}

Annotating datasets for object detection is an expensive and time-consuming endeavor. To minimize this burden, active learning (AL) techniques are employed to select the most informative samples for annotation within a constrained ``annotation budget". Traditional AL strategies typically rely on model uncertainty or sample diversity for query sampling, while more advanced methods have focused on developing AL-specific object detector architectures to enhance performance. However, these specialized approaches are not readily adaptable to different object detectors due to the significant engineering effort required for integration. To overcome this challenge, we introduce \ourmethodIntro, a simple and effective AL strategy for object detection. \ourmethodShort~is a two-stage method comprising uncertainty-based and diversity-based sampling phases. In the first stage, our \ourUncetaintySample~leverage a category-wise difficulty coefficient that combines both classification and localisation difficulties to re-weight instance uncertainties, from which we sample a candidate pool for the subsequent diversity-based sampling. In the second stage, we propose \ourSim~to better compute the similarities of multi-instance images as ensembles of their instance similarities, which is used by the \textit{k-Means++} algorithm to sample the final AL queries. \ourmethodShort~makes no change to model architectures or detector training pipelines; hence it can be easily generalized to different object detectors. We benchmark~\ourmethodShort~on the MS-COCO and Pascal VOC datasets using different detector architectures and show that our method outperforms prior work by a large margin. Code is available at \url{https://github.com/ChenhongyiYang/PPAL}

\end{abstract}

\section{Introduction}
\label{sec:intro}

\begin{figure}[!t]
\centering
    \includegraphics[width=\linewidth]{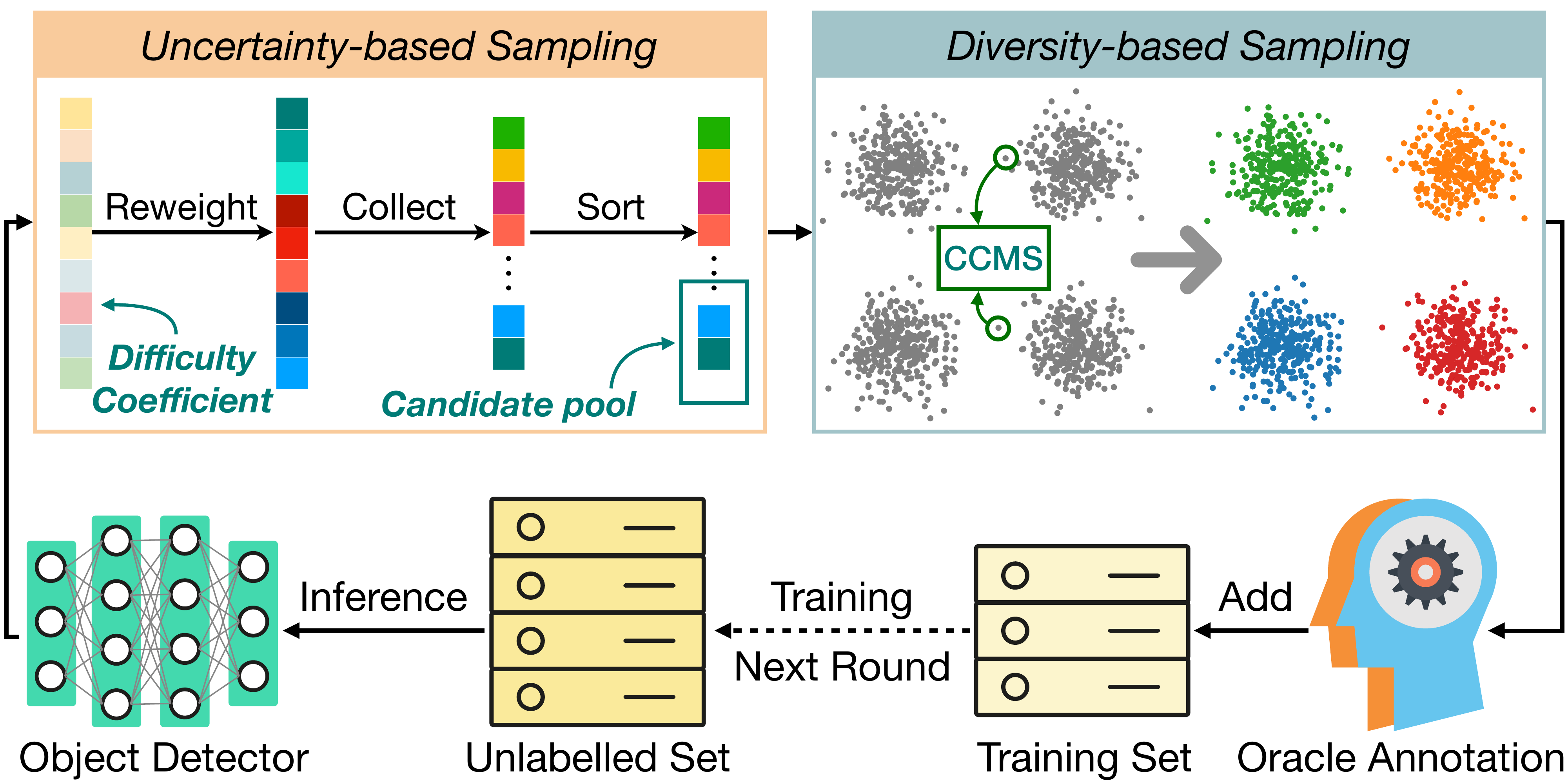}
    \caption{\small An overview of our two-stage \ourmethodShort. In the first \ourUncetaintySample~stage, the objects' uncertainties are re-weighted with the difficulty coefficients that take both classification and localisation into account, and a \textit{candidate pool} of images, which the model is mostly uncertain on, are sampled. In the second diversity-based stage, we run a modified kmeans++ algorithm using the proposed \ourSimFull~to select a set of representative images as active learning queries for the next round of annotation. }
    \label{fig:intro}
    \vspace{-0.5cm}
\end{figure}

Object detectors typically need a huge amount of training data~\cite{lin2014microsoft,kuznetsova2020open} annotated with both object category labels and bounding box locations. This annotation is expensive and tedious. If we are required to do some annotation, it would make sense to annotate the images that will be of the greatest benefit when used for training. But how do we know which ones to choose? The goal of active learning is to tell us. Given a large unlabelled data pool, active learning (AL) aims to sample data that would maximally improve a model's performance if that data was annotated and used for training. There are typically two main streams of active learning: (1) Uncertainty-based AL methods~\cite{settles2012active,lewis1994sequential,lewis1994heterogeneous,joshi2009multi,roth2006margin} select samples that maximise a measure of model uncertainty e.g.\ those with the least mutual information with the current set of labelled data; (2) Diversity-based AL approaches~\cite{sener2017active,agarwal2020contextual,luo2013latent,mac2014hierarchical,guo2010active,elhamifar2013convex,yang2015multi,bilgic2009link,wang2016cost} instead select samples that are representative of the whole distribution of unlabelled data; this can be achieved by minimising the similarities between the features~\cite{sener2017active} or posterior probabilities vectors~\cite{agarwal2020contextual} inside this subset.

With a neural network, uncertainty-based and diversity-based AL can be straightforwardly applied to the image classification task. For example, uncertainty-based sampling can be implemented by selecting the unlabelled samples with the maximum classification entropy, and the diversity-based can be achieved by minimising the selected samples' mutual similarities computed using their averaged feature maps. However, designing an effective AL strategy for object detection is more challenging. This is because detection consists of both object localisation and classification. It is difficult to quantify~\emph{uncertainty} jointly across both tasks e.g.\ we may have an object that is easy to localise, but hard to classify. Also, it is hard to measure~\emph{similarity} when images contain multiple objects with different features. 
Several works have tackled AL for detection~\cite{cai2020learning,yuan2021multiple,choi2021active} but rely on modifying the architecture of an object detector as well as the training pipeline. This means they cannot be easily integrated into other object detector frameworks without significant engineering effort.

In this work we propose \ourmethodIntro, a plug-and-play AL algorithm for object detection. It is state-of-the-art and easy to use; it requires no modifications to architectures or training pipelines and works across a wide range of object detectors. \ourmethodShort~is a two-stage algorithm that combines uncertainty- and diversity-based sampling: in the first stage it selects a \textit{candidate pool} of images with high uncertainty scores from a large unlabelled dataset, then in the second stage it select a highly diverse AL query set for annotation. In more detail, we propose \ourUncetaintySampleFull~for the first stage, in which the category-wise difficulty coefficients are computed and updated during training and then used to re-weight the instance uncertainties when selecting samples. The difficulty coefficients take both classification and localisation difficulties into account, so the two sub-tasks of object detection are balanced in the uncertainty-based sampling stage. \ourUncetaintySampleShort~also allows the uncertainty sampling to favour objects in challenging categories, hence benefiting the overall average precision (AP). For the second stage, we propose~\ourSimFull: a new method for measuring similarities for multi-instance images, in which every object is matched to its most similar counterpart in the other image to compute instance-wise similarity, and the image-wise similarity is computed by ensembling the instance-wise similarities. Then, \ourSimShort~is used off-the-shelf by a modified kmeans++ algorithm to select a diverse and representative subset as the active learning queries. Notably, unlike the recently proposed~\cite{cai2020learning,yuan2021multiple,choi2021active}, \ourmethodShort~does not modify the model architecture or training pipeline of object detectors, thus it is highly generalisable to different types of detectors. 

To summaries, our contributions are: \textbf{(1)} We propose \ourmethodShort, a two-stage active learning algorithm for object detection that combines uncertainty-based and diversity-based sampling. It is plug-and-play, requiring no architectural modifications or any change to training pipelines. \textbf{(2)} We show that \ourmethodShort~outperforms previous object detection AL algorithms across multiple object detectors on the COCO and Pascal VOC datasets. We also show that our method can be easily generalised to different object detectors.

\vspace{-1mm}
\section{Related Work}
\label{sec:related}
\vspace{-1mm}

\noindent\textbf{Active Learning.}
Active learning approaches can be broadly separated into uncertainty-based and diversity-based methods. Uncertainty-based methods~\cite{settles2012active,lewis1994sequential,lewis1994heterogeneous,joshi2009multi,roth2006margin} aim to select samples for annotation that maximise some uncertainty measure; common measures include entropy~\cite{settles2012active,joshi2009multi} and the margin between largest two predicted class posterior probabilities~\cite{roth2006margin,joshi2009multi,citovsky2021batch}. In~\cite{gal2017deep} the model uncertainty is estimated by performing multiple forward passes with Monte Carlo Dropout. Learn Loss~\cite{yoo2019learning} uses a prediction of the loss as a measure of uncertainty. Diversity-based methods~\cite{sener2017active,agarwal2020contextual,luo2013latent,mac2014hierarchical,guo2010active,elhamifar2013convex,yang2015multi,bilgic2009link,wang2016cost,wu2021redal} for AL aim to select representative samples so that a small subset of data can describe the whole dataset. Core-set~\cite{sener2017active} uses a greedy k-centroid algorithm to select a small core set from the unlabelled pool and use Mixed Integer Programming to iteratively improve sample diversity. CDAL~\cite{agarwal2020contextual} utilises predicted probabilities to improve context diversity. Recently, several works have been proposed that combine uncertainty-based and diversity-based AL. In~\cite{zhdanov2019diverse,Prabhu_2021_ICCV}, the uncertainty and diversity are balanced by running the k-means algorithm on image features weighted by model uncertainty. In BADGE~\cite{ash2019deep} the uncertainty and diversity are balanced by a k-means++ algorithm seeding on the gradients of the model's last layer.
HAC~\cite{citovsky2021batch} first clusters the unlabelled samples and queries the most uncertain samples in each cluster in a round-robin way.

\noindent\textbf{Object Detection.}
ConvNet-based object detectors usually follow a two-stage or single-stage design. Two-stage detectors~\cite{ren2015faster,he2017mask,cai2018cascade,chen2019hybrid} use a region proposal network~\cite{ren2015faster} to extract plausible objects and the RoIAlign operation~\cite{he2017mask} to extract regional features for further classification and localisation; Single-stage~\cite{huang2015densebox,redmon2018yolov3,liu2016ssd,chen2021ddod,lin2017focal,tian2019fcos,zhang2020bridging,rezatofighi2019generalized} detectors directly predict objects' bounding box and category label on every position of the image feature maps. Recently, a wave of transformer-based detectors~\cite{carion2020end,zhu2020deformable,li2022dn,zhang2022dino} have been proposed. They use self-attention to exchange information between query vectors and image features and directly output detected objects without Non-maximum-suppression.

\noindent\textbf{Active Learning for Object Detection.} Early attempts at applying AL to object detection~\cite{agarwal2020contextual,sener2017active,yoo2019learning,Yu_2022_CVPR,lyu2023box} involved the direct application of image classification AL algorithms. However, these do not account for the joint classification and localisation tasks that make up detection, or that images can contain multiple different objects. This prompted the design of AL algorithms specifically for detection. MDN~\cite{choi2021active} modify an object detector to learn a Gaussian mixture model (GMM) for both the classification and regression outputs, and then the uncertainties of both tasks are derived from the modelled GMMs. In~\cite{elezi2022not}, semi-supervised learning is used to deal with false positives and compute model uncertainty. MIAL~\cite{yuan2021multiple} uses adversarial training to compute model discrepancy, which is used for computing uncertainty. DivProto~\cite{wu2022entropy} improves the AL algorithm by replacing the detection scores with model uncertainties in NMS; it also uses a set of diverse prototypes to select the most representative samples. However, a problem of previous works in AL for object detection is that the model architecture or training pipelines are modified to suit the AL purpose, which limits their generalisation ability across different architectures.

\vspace{-1mm}
\section{Method}
\label{sec:method}
\vspace{-1mm}

In this section, we first define the problem of active learning for object detection (Sec.~\ref{sec:method_problem}). Then, the two key innovations of our method are described in detail in Sec.~\ref{sec:uncertainty} and~\ref{sec:method_diversity}. Fig.~\ref{fig:intro} gives a high-level overview of our algorithm.

\subsection{Problem Statement}
\label{sec:method_problem}

Following~\cite{choi2021active, yuan2021multiple, wu2022entropy}, we define the problem under the batch active learning setting, i.e.\ at each round we query a batch of images for oracle annotation instead of a single sample. Suppose there is a training set $X_T=\{x_i\}_{i\in [N_t]}$ with size $N_t$ and a validation set $X_V=\{x_i\}_{i\in [N_v]}$ with size $N_v$. At round $r \geq 0$, the labelled set is $X_L^r$ and the unlabelled pool is $X_U^r$ where $X_L^r \cap X_U^r = \O$ and $X_L^r \cup X_U^r = X_T$. An object detection model $f^r_{\theta}$ with parameters $\theta$ is trained on $X_L^r$ and its performance on $X_V$ is $Z(X_V|f^r_\theta)$, which is usually measured by  mean Averaged Precision (mAP). Given budget $b$, an active learning algorithm selects a query set $X_Q^r \subseteq X_U^r$ with size $b$ for oracle annotation. Then we can get the labelled set  $X_L^{r+1} = X_L^{i} \cup X_Q^r$ and the unlabelled  $X_U^{r+1} = X_U^{i} - X_Q^r$ for the next round $r+1$. Finally, the detection model is trained on $X_L^{r+1}$ to get $f^{r+1}_{\theta}$ with performance $Z(X_V|f^{r+1}_\theta)$. After $k$ rounds of active learning, an active learning algorithm is evaluated by the model's improvement over the initial round $\triangle Z^{k,0}= Z(X_V|f^{k}_\theta)-Z(X_V|f^{0}_\theta)$, i.e., the AL algorithm that can bring most performance improvement is favoured. 
\begin{figure}[!t]
\centering
    \includegraphics[width=\linewidth]{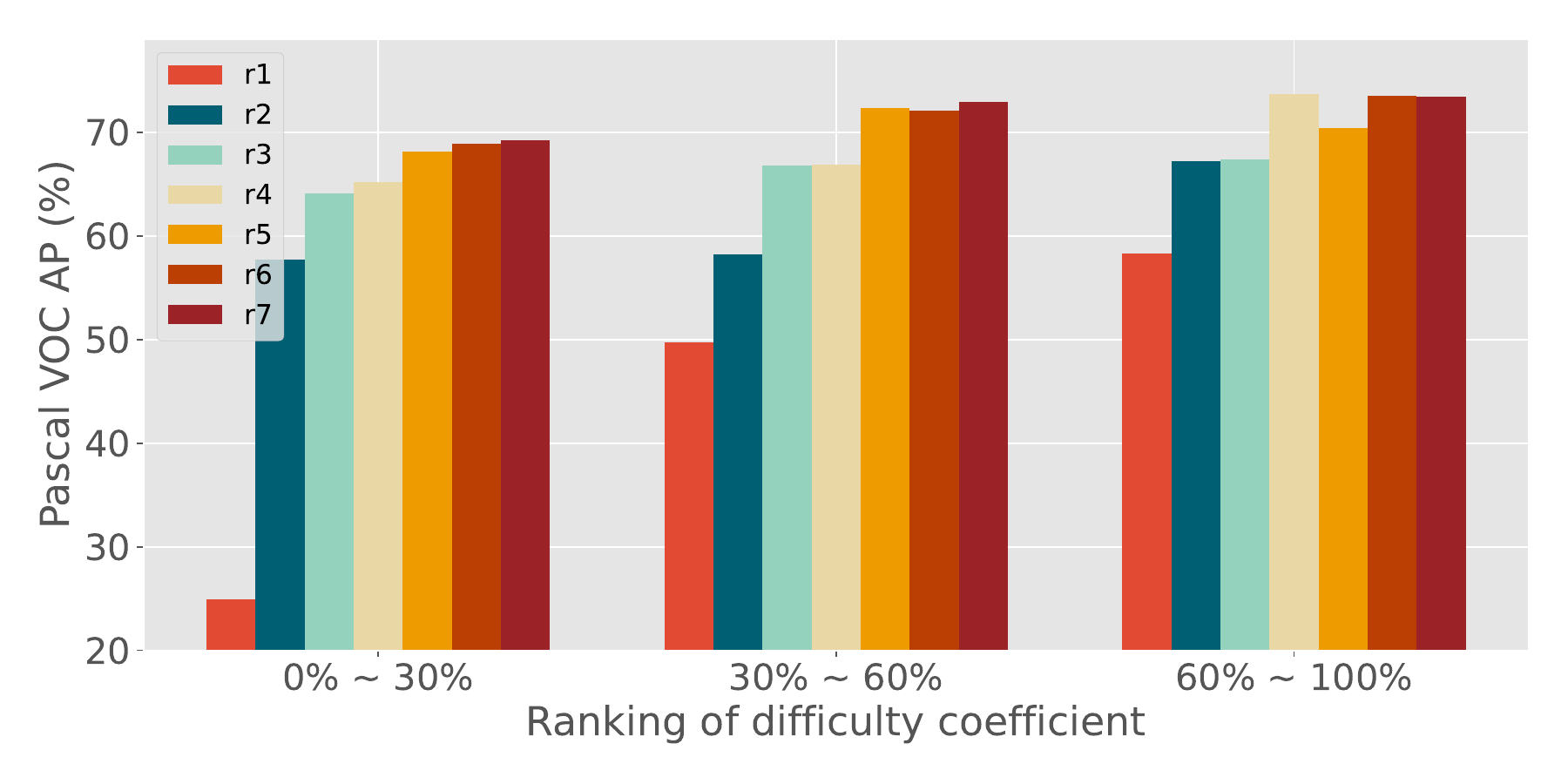}
    \vspace{-8mm}
    \caption{\small Illustration of how the category-wise difficulty coefficients correspond to the evaluated detection APs on Pascal VOC at each active learning round, in which the difficulty coefficients are sorted in descending order. Objects in categories with high-difficulty coefficients are harder to be detected than those in categories with low-difficulty coefficients.}
    \label{fig:coeffcient}
    \vspace{-4mm}
\end{figure}

\subsection{\ourUncetaintySample}
\label{sec:uncertainty}

We propose \ourUncetaintySampleFull~to serve two purposes: (1) it provides a means to score uncertainty for object detection based on both classification and localization; (2) it allows more objects in challenging categories to be sampled. \ourUncetaintySampleShort~circumstance the above two challenges by re-weighting the object uncertainties with a category-dependent difficulty coefficient. Intuitively, we aim to raise the importance of categories that the model does not perform well, while down-weight the easy categories. Specifically, suppose there are $C$ classes in the dataset, at the beginning of each AL training round, we define the class-wise difficulties $D^r = \{d_i\}_{i\in [C]}$ and initialise them with all 1. Inspired by~\cite{chen2021ddod}, we compute the training difficulty of every predicted box during training as:
\begin{equation}
\begin{small}
\label{eq:difficulty}
	\begin{aligned}
        q(b | \hat{b}) = 1- {P\bigl(b|\hat{b}\bigr)}^{\xi} \cdot { \mathrm{IoU} \bigl( b, \hat{b} \bigr) }^{1-\xi}
    \end{aligned}
\end{small}
\end{equation}
where $b$ is the predicted box, $\hat{b}$ is the ground-truth box that $b$ is assigned to, $P\bigl(b|\hat{b}\bigr)$ is the classification probability w.r.t. the class of $\hat{b}$, $\mathrm{IoU} \bigl( b, \hat{b} \bigr)$ is the IoU between $b$ and $\hat{b}$, and $0\leq \xi \leq 1$ is a hyper-parameter. The detection difficulty defined in Eq.~\ref{eq:difficulty} takes both classification and localisation into account, hence both of them will contribute to the uncertainty sampling. Then, the recorded class-wise difficulties are updated by the averaged object difficulty using an exponential moving average (EMA) during training:
\begin{align}
        &d_i^k 	\leftarrow m_i^{k-1}d_i^{k-1} + (1-m_i^{k-1}) \frac{1}{N^k_i} \sum_{j=1}^{N^k_i}{q_j}\label{eq:diffucltyUpdate} \\
        &m_i^{k} \leftarrow  
        \begin{cases}
        m^0 \text{~~~~~~~~~~~~~~~~~~~~~~~~~~if $N_i^k > 0$} \\
        m^0 \cdot m_i^{k-1} \text{~~~~~~~~~~~~~~if $N_i^k = 0$}\label{eq:momentumUpdate}
      \end{cases}
\end{align}where $k$ is the training iteration, $d_i^k$ is the updated difficulty for category $i$, $N^k_i$ is the number of predicted objects of class $i$ in the training batch, $q_j$ is the $j$-th object's training difficulty, $m_i^*$ is the EMA momentum, $m^0$ is the initial momentum for all categories. As shown in Eq.~\ref{eq:momentumUpdate}, if a training batch does not contain objects in category $i$, we decrease its class-wise momentum by multiplying $m^0$, which ensures a similar updating pace of difficulties of different classes. In Fig.~\ref{fig:coeffcient}, we show how the class-wise difficulties correspond to the class-wise detection APs during active learning.

\begin{figure}[!t]
\centering
    \includegraphics[width=\linewidth]{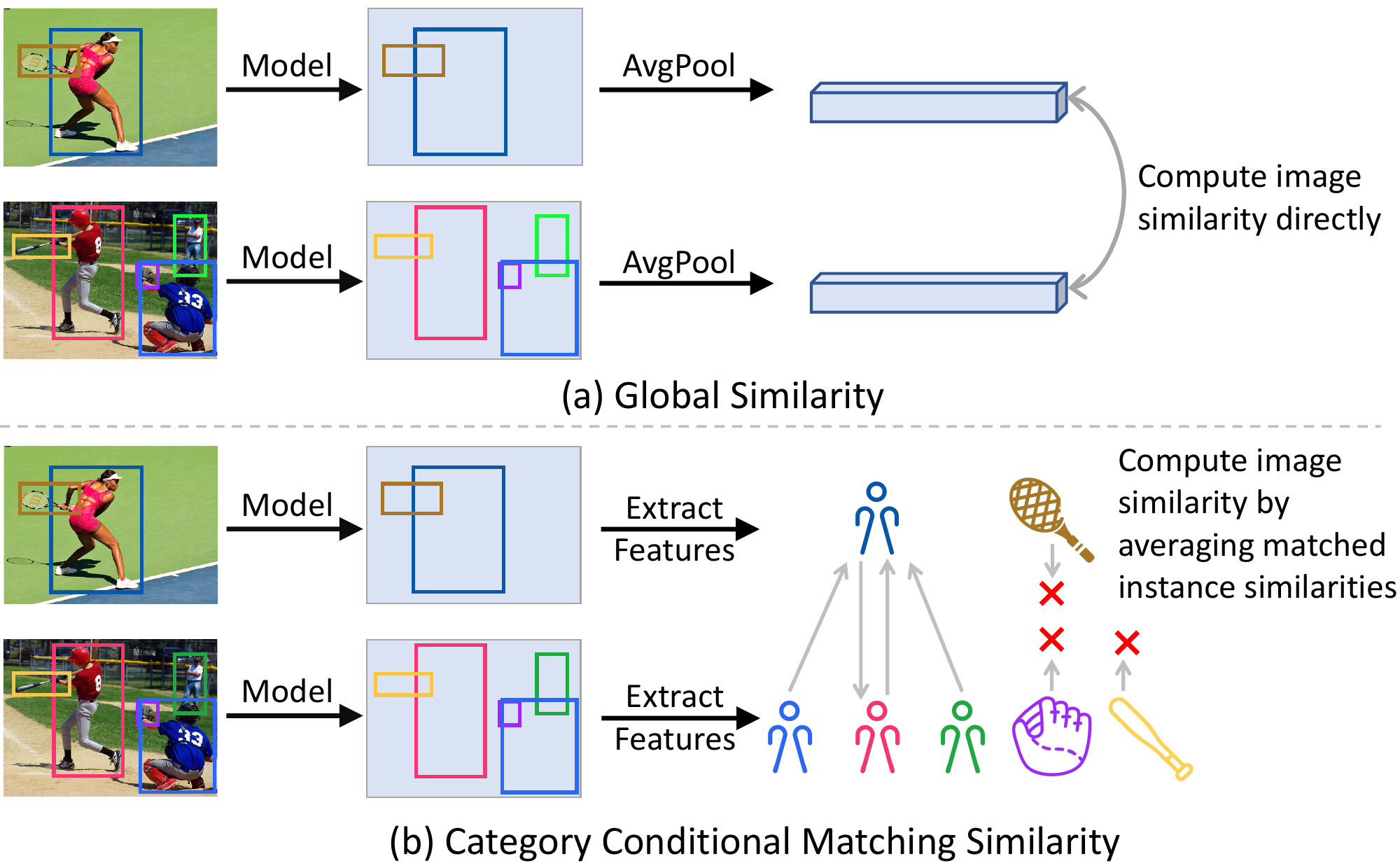}
    \caption{\small Comparison of global similarity and our \ourSimShort. The global similarity is computed using the averaged image feature maps, failing to capture the fine-grained spatial information of multi-instance images. On the other hand, in \ourSimShort, each object in an image finds its most similar counterpart with the same category in another image to compute similarities. Then image-wise similarity is computed by averaging the object similarities.}
    \label{fig:sim}
    \vspace{-0.4cm}
\end{figure}

After training the detector on the labelled dataset, we compute the category-wise difficulty coefficient $W^r=\{w_i\}_{i\in [C]}$ for the $r$-th AL round as:
\begin{equation}
\label{eq:coefficient}
	\begin{aligned}
        &w_i = 1 + \alpha \beta \cdot \log{(1 + \gamma \cdot d_i)}\\ 
        &\text{where~~~~}  \gamma = e^{1 / \alpha} - 1
    \end{aligned}
\end{equation}
where $\alpha$ controls how fast the difficulty coefficient changes w.r.t. the class-wise difficulty, $\beta$ controls the upper bound of the difficulty coefficient. Finally, we compute the image-wise uncertainty of every unlabelled image by summing the entropy of each detected object weighted by the corresponding difficulty coefficient:
\begin{align}
\begin{small}
    U(I) = \sum_{i=1}^{M_{I}} w_{c(i)} \cdot \sum_{j=1}^{C'}{-p_{ij} \cdot \log(p_{ij})}
\end{small}
\end{align}where $M_{I}$ is the number of detected objects from image $I$; $w_{c(i)}$ is the weight of object $i$'s predicted category; $C'$ is the number of classification ways, which is usually $C+1$ for two-stage detectors and $2$ for one-stage detectors; $p_{ij}$ is the predicted probability of category $j$. Finally, we use a simple strategy to select the candidate pool: for a given AL budget $b$, we sort the images by their uncertainties and select $\delta \cdot b$ most uncertain ones from the unlabelled set. We call the hyper-parameter $\delta > 1$ \textit{budget expanding ratio}. 

\subsection{Diversity Sampling for Multi-instance Images}
\label{sec:method_diversity}

In the second stage of \ourmethodShort~a diverse and representative set will be selected from the candidate pool to serve as the AL query set. In previous works~\cite{sener2017active,agarwal2020contextual}, diversity-based sampling is achieved by minimising the similarities of every pair of the selected samples. Such similarities are often computed by the cosine or $L_2$ similarities of the averaged convolutional feature maps~\cite{sener2017active}, which we call \textit{global similarity}. This practice is simple and works well for object-centric datasets like ImageNet~\cite{imagenet}. However, object detection usually take multi-instance images as input, and in such cases the averaged feature maps are difficult to capture the fine-grained spatial information in those images~\cite{yang2021contrastive}. Therefore, we design a new similarity computing method to compute similarities of multi-instance images, which can be used off-the-shelf for diversity-based sampling. We show the differences between those two similarities in Fig.~\ref{fig:sim}.

\begin{figure*}[!ht]
    \centering
    \includegraphics[width=0.9\textwidth]{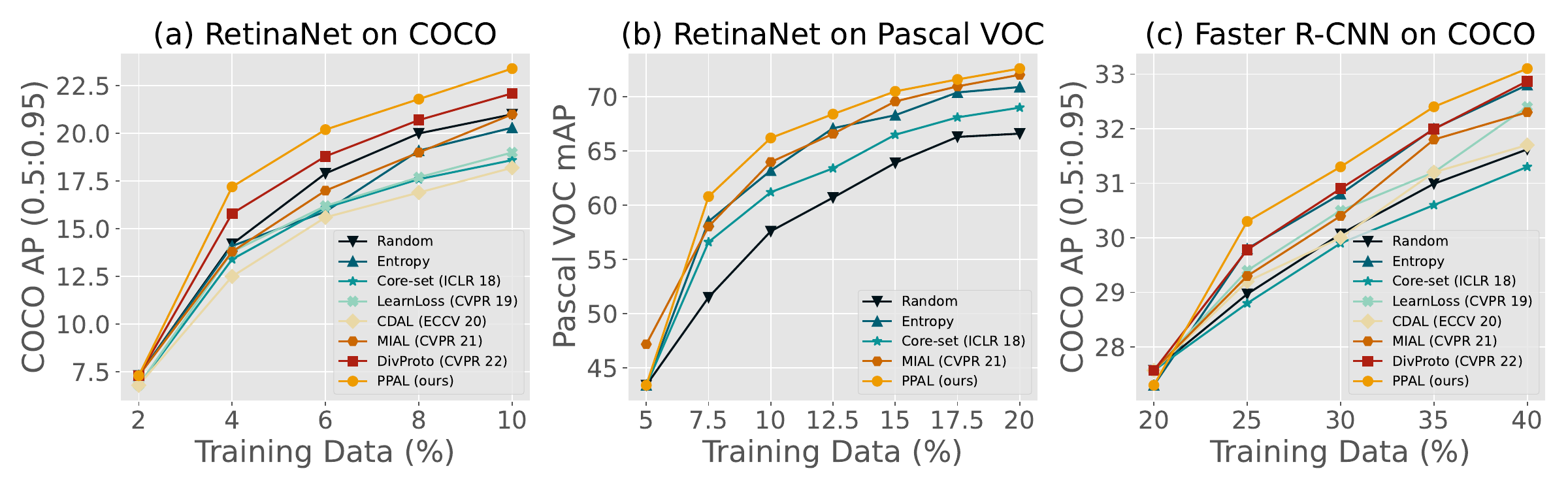}
    \vspace{-5mm}
    \caption{\small Comparison between the proposed method and the state-of-the-art active learning algorithm for object detection in three different benchmark settings. (a) RetinaNet on COCO; (b) RetinaNet of Pascal VOC; (c) Faster R-CNN on COCO.}
    \label{fig:sota}
    \vspace{-0.4cm}
\end{figure*}

\noindent \textbf{\ourSim.} The intuition behind our \ourSimShort~is that the similarity of two multi-instance images can be computed by measuring how similar their contained objects are. Formally, for two multi-instance images $I_a$ and $I_b$, the object detector detects several objects from them $O_a=\{o_{a,i}\}_{i\in[M_a]}$ and  $O_b=\{o_{b,i}\}_{i\in[M_b]}$, in which each object $o_{*,i}$ is a triplet $o_{*,i}=(f_{*,i},t_{*,i}, c_{*,i})$: $f_{*,i}$ is the object's visual features extracted from the feature maps; $t_{*,i}$ is the detection score; $c_{*,i}$ is the predicted class label. We use the similarity of $O_a$ and $O_b$ as a proxy for the similarity of $I_a$ and $I_b$, which is computed by matching every object to its most similar counterpart in the same category in the other image. Specifically, for an object $o_{a,i}$ in $O_a$, its similarity to $O_b$ is computed as: 
\begin{equation}
    s(o_{a,i}, O_b) =
      \begin{cases}
        \underset{c_{b,j}=c_{a,i}}{\max} \frac{f_{a,i} \cdot f_{b,j}}{||f_{a,i}|| \cdot ||f_{b,j}||} + 1 \\
        ~~~~~0~~~~~ \text{if no $c_{b,j}=c_{a,i}$}
      \end{cases}
\end{equation} where we set $S(o_{a,i}, O_b)$ to $0$ if no object in $O_b$ is in the same category as $o_{a,i}$. Then the similarity of $O_a$ to $O_b$ is computed by averaging the object similarities weighted by their detection scores:
\begin{equation}
\begin{aligned}
    &S'(O_a, O_b) = \frac{1}{\sum_i{t_{a,i}}}\sum_{i=1}^{M_a}{t_{a, i} \cdot s(o_{a,i}, O_b)} \\
    &S(O_a, O_b) = \frac{1}{2} \cdot \bigl(S'(O_a, O_b) + S'(O_b, O_a)\bigr)
\end{aligned}
\end{equation}where the final similarity is the average of $S'(O_a, O_b) $ and $S'(O_b, O_a)$, which ensures the symmetry of the similarity. 

\begin{figure*}[!ht]
    \centering
    \includegraphics[width=0.9\textwidth]{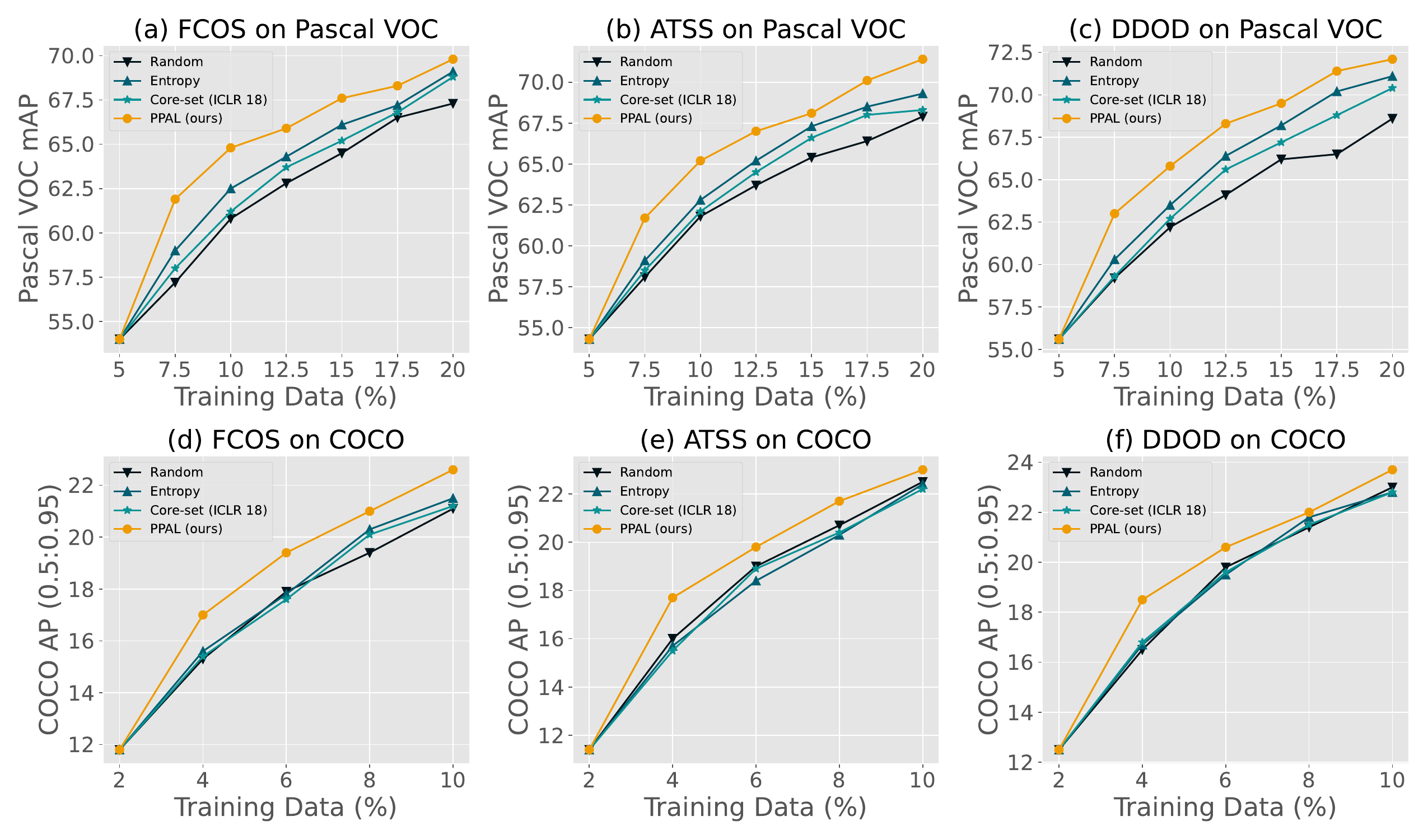}
    \vspace{-0.4cm}
    \caption{\small Active learning on Pascal VOC and COCO using (a) Anchor-free FCOS; (b) Anchor-based ATSS; (c) Anchor-based DDOD.}
    \label{fig:diff}
    \vspace{-0.4cm}
\end{figure*}

\noindent \textbf{Sampling AL Queries.} We use the proposed \ourSimShort~to sample representative image set from the candidate pool as the AL query $Q$. The objective of the diversity-based sampling is formally written as: 
\begin{align}
\label{eq:diversity}
    Q = \underset{Q' \subseteq H: |Q'|=b}{\min} \Bigl\{ \underset{I_i, I_j \in Q'}{\max} S(O_i, O_j)\Bigr\}
\end{align}
where $H$ is the candidate pool output from the first stage, and $b$ is the AL budget. However, Eq.~\ref{eq:diversity} is an NP-Hard problem~\cite{sener2017active}, so we follow~\cite{sener2017active} to use the k-Center-Greedy algorithm to get a $2-OPT$ solution. Another problem of the objective in Eq.~\ref{eq:diversity} is that it only maximises the diversity of the selected samples while ignoring how representative they are, which may cause the selection to favour data outliers. To deal with the problem, we further run a k-means++ algorithm using the results of k-Center-Greedy algorithm as the initial centroids. However, here we can only compute the similarities of every pair of images and are not able to compute the actual \textit{mean} of every cluster to update its centroids. We solve this problem by assigning the new centroids as the images whose summed similarities to other images in their cluster are maximised. Finally, the resulting AL queries are sent to human experts for annotation.

\vspace{-3mm}
\section{Experiments}
\label{sec:exp}
\vspace{-1mm}

\subsection{Experiment Settings}
\vspace{-1mm}
\noindent\textbf{Dataset settings.}
We benchmark the proposed \ourmethodShort~using two datasets: COCO~\cite{lin2014microsoft} and Pascal VOC~\cite{everingham2015pascal}. For COCO we use \textit{train2017} set for training, and evaluate the models on the \textit{mini-val} set. For Pascal VOC, we use \textit{train2007+2012} for training and \textit{test2007} for testing. When comparing with previous works, we follow their dataset split settings~\cite{yuan2021multiple,wu2022entropy} to ensure fair comparisons. Note that those settings may vary across different detectors. Ablation studies are conducted on Pascal VOC with a unified setting: 5\% of the training data are first sampled as the initial set, then 6 rounds of active learning are conducted where at each round 2.5\% extra data are queried. To overcome randomness, we run all experiments using three different initial training sets and report the averaged performance. 

\noindent\textbf{Model settings.}
By default, we follow~\cite{yuan2021multiple} to set our model and training recipes, which ensures a fair comparison of our method and previous works. We implement our code base using the MMDetection toolkit~\cite{mmdetection}. For both dataset we train the models for 26 epochs and decay the learning rate by 0.1 at the 20th epoch. We use ResNet-50~\cite{he2016deep} as the default backbone network. All experiments are conducted using 8 2080Ti NVIDIA GPUs. We follow~\cite{chen2021ddod} to set the $\xi$ in Eq.\ref{eq:difficulty} to 0.6; we set base EMA momentum $m^0$ to to 0.99 following common practices. Other hyper-parameters are empirically set without careful tunning: $\alpha$ and $\beta$ in Eq.\ref{eq:coefficient} are set to 0.3 and 0.2, and the budget expansion ratio $\delta$ is set to 4. We run the kmeans++ algorithm for 100 iterations.

\begin{figure}[!t]
\centering
    \includegraphics[width=\linewidth]{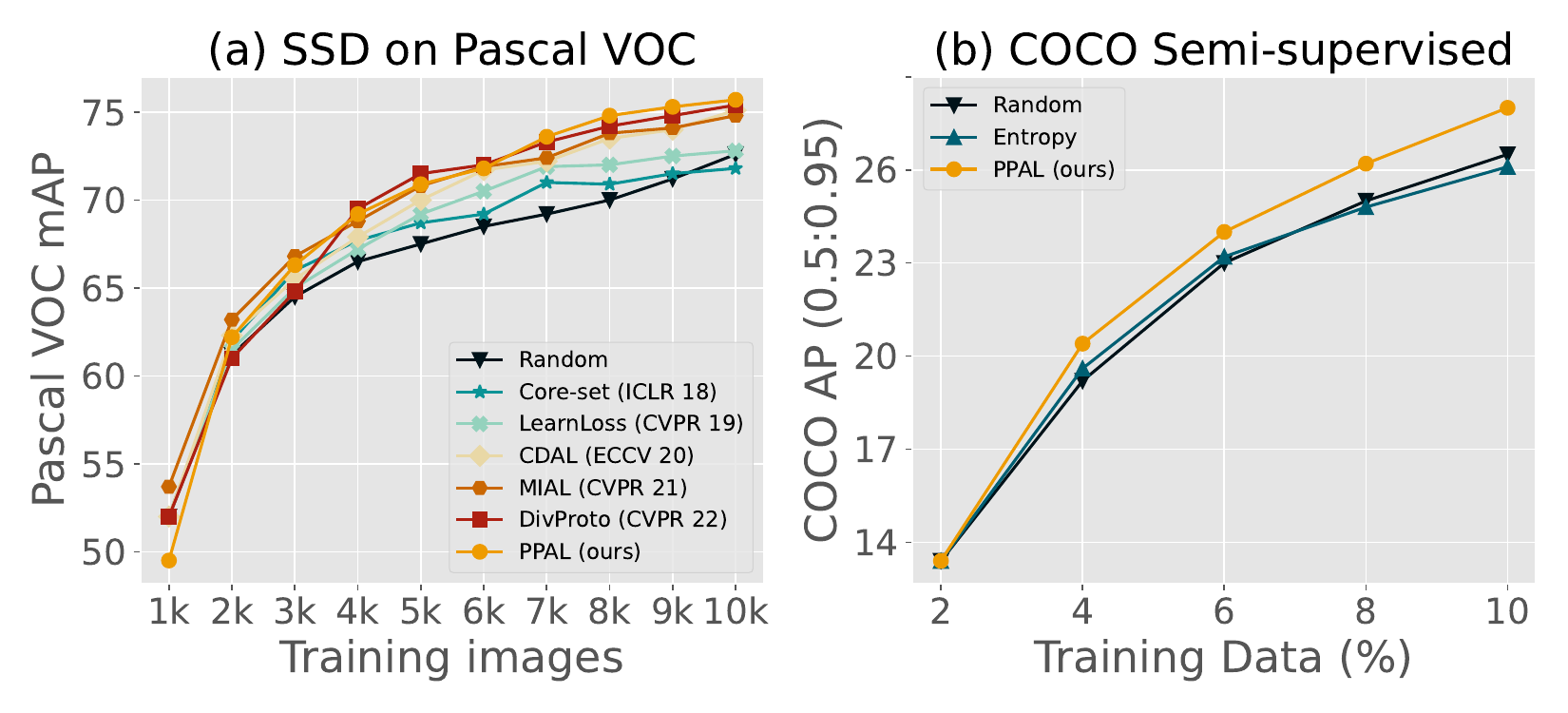}
    \vspace{-8mm}
    \caption{\small (a) Comparison between the proposed method and the state-of-the-art active learning algorithm for object detection on Pascal VOC with SSD object detector. (b) Active Learning performance on COCO using semi-supervised Soft Teacher detector.}
    \label{fig:ssdSemi}
    \vspace{-5mm}
\end{figure}

\subsection{Main Results}

\noindent\textbf{Comparing with the state-of-the-art.}
In Fig.\ref{fig:sota}, we compare our \ourmethodShort~with previous state-of-art active learning strategies on three different settings: COCO RetinaNet~\cite{yuan2021multiple}, Pascal VOC RetinaNet~\cite{yuan2021multiple}, and COCO Faster R-CNN~\cite{wu2022entropy}.  We follow the training and dataset settings in~\cite{yuan2021multiple} for the first two comparisons and follow~\cite{wu2022entropy} for the last. For the entropy-based sampling baseline, we sum up the classification entropy of all detected objects as the image uncertainty, which results in a better performance than previous works~\cite{elezi2022not,yuan2021multiple,choi2021active} that compute image-wise uncertainty using the averaged or the maximum object entropy. As shown by the results, our \ourmethodShort~outperforms all computing methods in all three settings. Specifically, in the initial rounds ($\leq$ 3), our method outperforms all competing methods by large margins, e.g., it outperforms MIAL~\cite{yuan2021multiple} by 3.4 AP and 2.8 mAP on COCO and Pascal VOC respectively, which suggests that \ourmethodShort~can provide the detectors with highly informative samples when they are not well trained. In the later rounds ($>$ 3), although the AP gaps between \ourmethodShort~and the competing methods decrease, it still maintains the lead. In addition to \ourmethodShort's ability to mine informative data samples, it also owns a better generalisation ability than the competing methods~\cite{yuan2021multiple, yoo2019learning}, because it does not modify the model architecture or training pipeline. 

\noindent\textbf{Performance on more detectors.}
Our \ourmethodShort~is highly generalisable, which allows us to easily apply it to different object detectors. In Fig.\ref{fig:diff}, we show the Pascal VOC and COCO results when applying it to three recently proposed object detectors: FCOS~\cite{tian2019fcos}, ATSS~\cite{zhang2020bridging} and DDOD~\cite{chen2021ddod}, in which FCOS is anchor-free, and the rest are anchor-based. We also shows the results of three baselines: \textit{Random}, \textit{Entropy} (uncertainty-based), and Core-set~\cite{sener2017active} (diversity-based). The results show that \ourmethodShort~is consistently better than the baselines for all three detectors, verifying its effectiveness and good generalisation ability.

\noindent\textbf{SSD Experiments.}
In Fig.~\ref{fig:ssdSemi} (a), we compare \ourmethodShort~with previous works on Pascal VOC using the SSD~\cite{liu2016ssd} detector. We follow the training recipes in MIAL~\cite{yuan2021multiple} to train the model for 300 epochs at each active learning round. Unlike other detectors~\cite{ren2015faster,lin2017focal} that were used in the main paper, in SSD objects detected from different feature levels are computed using different convolutional kernels. In this case, we are unable to compute those objects' distances using their visual features, so we follow CDAL~\cite{agarwal2020contextual} to compute the distances of their classification probability vectors using KL-divergence. The results show that \ourmethodShort~is able to achieve a better performance than all competing methods. Notably, although the initial performance of our SSD model is inferior to others, our active learning approach can still enable the model to surpass others in the later stages. 

\noindent\textbf{Semi-supervised Experiments.}
In Fig.~\ref{fig:ssdSemi} (b), we compare PPAL with \textit{Random} / \textit{Entropy} baselines using the semi-supervised detector SoftTeacher~\cite{Xu_2021_ICCV}. We followed the experiment settings in {Sec.4.1} to run five rounds of active learning on COCO. In each round, the model was trained for 26 epochs (counted using labelled images), and the class difficulties were computed using the labelled images. We observe that \textit{Entropy} active learning strategy achieves similar performance with the \textit{Random} baseline, but our PPAL is better than both of them. The result validates PPAL's effectiveness in a semi-supervised learning setting.

\subsection{Ablation Studies and Discussions}

\begin{table}[t]
  \begin{center}
  \resizebox{\linewidth}{!}{
    \begin{tabular}{c|c|cccccc}
    \hline
    \multirow{2}{*}{Stage 1} & \multirow{2}{*}{Stage 2} & \multicolumn{6}{c}{mAP on \% of labelled images} \\
    \cline{3-8}
      &  & 7.5\% & 10\% & 12.5\% & 15\% & 17.5\% & 20\%\\
    \hline
    \multicolumn{2}{c|}{Random} & 51.5$\pm$1.5 & 57.6$\pm$1.6 & 60.7$\pm$1.3 & 63.9$\pm$1.0 & 66.3$\pm$1.1 & 66.6$\pm$1.0 \\
    \hline
    Entropy & None & 58.5$\pm$0.7 &  63.2$\pm$0.6 &  67.1$\pm$0.3 & 68.3$\pm$0.3 & 70.4$\pm$0.3 & 70.9$\pm$0.2 \\
    \ourUncetaintySampleShort & None & 60.5$\pm$0.9 & 64.4$\pm$0.6 & 67.2$\pm$0.7 & 68.7$\pm$0.3 & 70.4$\pm$0.2 & 71.6$\pm$0.2 \\ %
    \hline
    Rand & \ourSimShort & 56.6$\pm$1.0 &  61.2$\pm$0.5 &  64.7$\pm$0.7 & 66.9$\pm$0.4 & 68.4$\pm$0.3 & 70.3$\pm$0.2 \\
    Entropy & \ourSimShort & 59.0$\pm$0.6 & 64.5$\pm$0.6 & 67.6$\pm$0.4 & 69.2$\pm$0.4 & 71.1$\pm$0.3 & 72.0$\pm$0.3  \\
    D-Freq & \ourSimShort & 59.3$\pm$0.6 & 64.2$\pm$0.5 & 67.9$\pm$0.5 & 68.8$\pm$0.5 & 71.4$\pm$0.3 & 71.8$\pm$0.3  \\
    \hline
    \ourUncetaintySampleShort & Rand  & 60.3$\pm$0.7 & 64.1$\pm$0.5 & 67.6$\pm$0.5 & 68.9$\pm$0.4 & 70.5$\pm$0.3 & 72.0$\pm$0.2 \\ %
    \ourUncetaintySampleShort & Global  & 58.8$\pm$0.6 & 64.5$\pm$0.4 & 66.9$\pm$0.2 & 68.6$\pm$0.2 & 69.3$\pm$0.4 & 71.8$\pm$0.2  \\ 
    \ourUncetaintySampleShort & FPN & 59.4$\pm$0.6 & 64.3$\pm$0.3 & 67.3$\pm$0.3 & 68.6$\pm$0.2 & 70.0$\pm$0.1 & 71.6$\pm$0.1\\ 
    \ourUncetaintySampleShort & Jaccard & 60.3$\pm$0.4 & 65.1$\pm$0.4 & 67.3$\pm$0.4 & 68.6$\pm$0.3 & 70.5$\pm$0.3 & 71.9$\pm$0.2\\ 
    \hline
    \ourUncetaintySampleShort & \ourSimShort & 60.8$\pm$0.5& 66.2$\pm$0.4 & 68.4$\pm$0.2 & 70.5$\pm$0.4 & 71.6$\pm$0.3 & 72.6$\pm$0.2  \\
    \hline
    \end{tabular}}
  \end{center}
  \vspace{-0.6cm}
  \caption{\small Ablation studies of \ourUncetaintySampleFull~and \ourSimFull~using VOC RetinaNet. The 1st round mAP is 43.4$\pm$2.2.}
  \vspace{-0.4cm}
\label{tbl:ablation}
\end{table}

\begin{figure}[!t]
\centering
    \includegraphics[width=\linewidth]{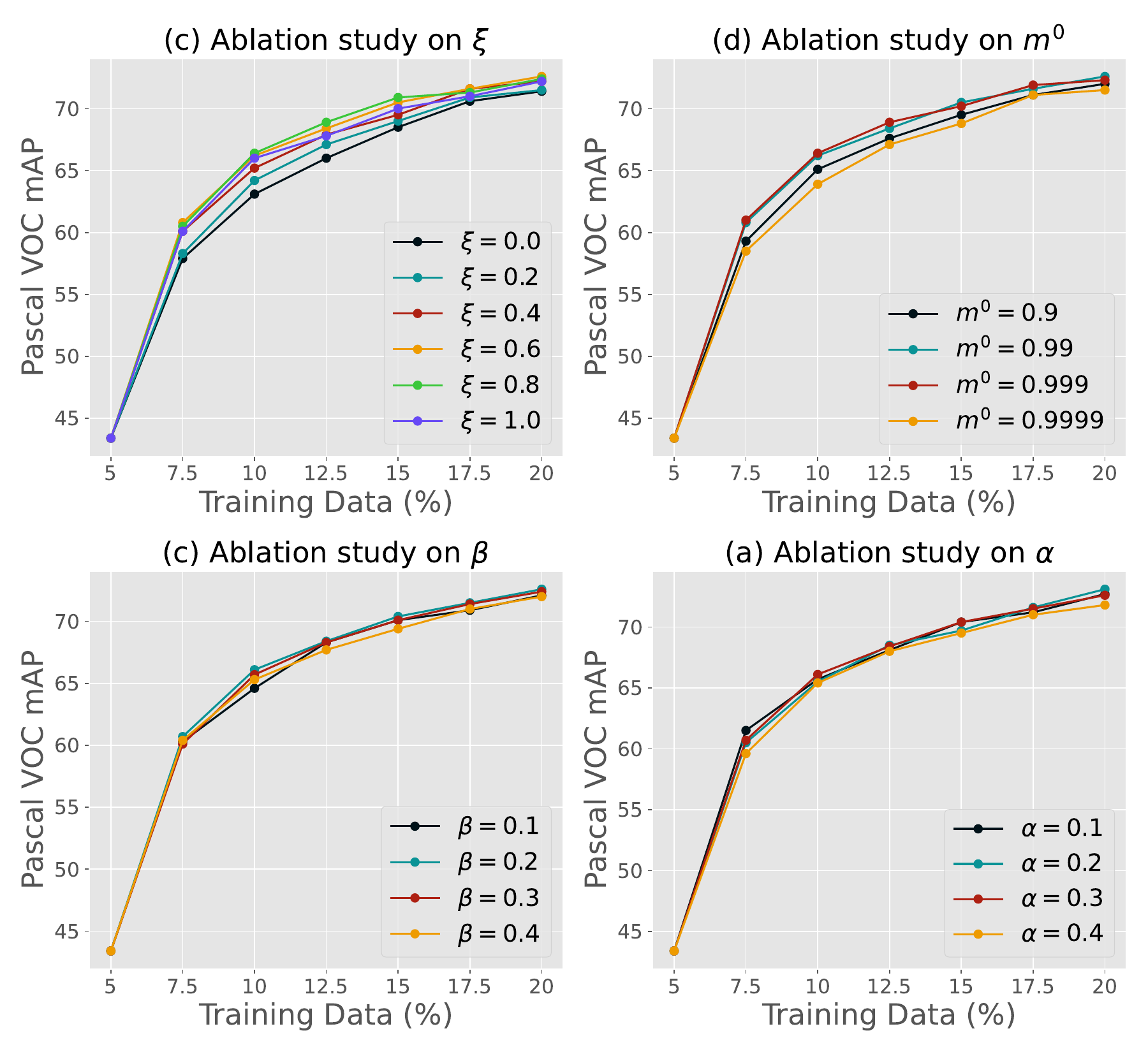}
    \vspace{-8mm}
    \caption{\small Ablation study on four hyper-parameter ablation studies in the proposed \ourUncetaintySampleShort~using Pascal VOC RetinaNet.}
    \label{fig:allHyperparameter}
    \vspace{-4mm}
\end{figure}

\noindent\textbf{Effectiveness of \ourmethodShort.}
As shown in Tab.\ref{tbl:ablation}, we start by comparing \ourUncetaintySampleShort~with random sampling and entropy-based sampling by using those strategies to sample the candidate pool and run the diversity-based sampling using our \ourSimShort~for similarity computing. We find that \ourUncetaintySampleShort~ achieves a better performance, which validates \ourUncetaintySampleShort's effectiveness as an uncertainty-based AL strategy. We also try to replace the difficulty coefficient described in Eq.~\ref{eq:coefficient} with $1-f_c$ where $f_c$ is class $c$'s frequency in the training set, which is denoted as \textit{DCUS-Freq}. Therefore the uncertainty-based sampling will favour those 
\textit{long-tailed} classes. The result shows this strategy is inferior to our \ourUncetaintySampleShort. A possible reason is that we find the class-wise AP does not strictly correspond to the training class frequencies. Therefore, in our proposed \ourUncetaintySampleShort, we turn to explicitly modelling the class-wise detection difficulties, which is more suitable for uncertainty re-weighting. Also in Tab.\ref{tbl:ablation}, we compare the proposed \ourSimShort~with four alternatives: (1) \textit{Rand}, using random selection for the second stage; (2) \textit{Global}, global similarity computed as the cosine similarity of the averaged feature maps from the backbone network's last layer; (3) \textit{FPN}, the averaged global similarity of every feature pyramid layer; (4) \textit{Jaccard}, the Jaccard similarity between the predicted category sets of two images. Note that nearly all previous works~\cite{yuan2021multiple,sener2017active} use global similarity for diversity sampling. We observe that the global similarity works even worse than random selection, demonstrating its failure in measuring the similarity of multi-instance images. The FPN similarity works slightly better than the global similarity and generates similar performance with random sampling. Our \ourSimShort~achieves the best result, which is consistently better than the three alternatives. These benchmark results validate that the proposed \ourSimShort~is more suitable to measure the similarities of multi-instance images in active learning. In addition, we observe that when diversity-sampling is fixed or disabled, our \ourUncetaintySampleShort~can achieve much better performance than entropy-based baseline in early AL rounds, but in later rounds their difference is marginal. On the other hand, although \ourSimShort~achieves a similar performance with random sampling in the first two rounds, but their performance gap starts to increase from the third round. From those observations we draw an important conclusion: \textit{In active learning for object detection, uncertainty-based sampling is more critical in the early AL stages while diversity-based sampling is more essential in later rounds.}

\noindent\textbf{Hyper-parameters.}
In Fig.\ref{fig:allHyperparameter}, we report the ablation study results to investigate the hyper-parameter settings in \ourmethodShort. They are $\xi$ in Eq.~\ref{eq:difficulty}, $m^0$ in Eq.~\ref{eq:momentumUpdate}, and $\alpha$ and $\beta$ in Eq.~\ref{eq:coefficient}. The results suggest that \ourmethodShort's performance is stable although some optimal hyper-parameters settings may exist.

\begin{table}[t]
  \begin{center}
    \resizebox{\linewidth}{!}{
    \begin{tabular}{c|cccccc}
    \hline
    {Pool Size}  &\multicolumn{6}{c}{\small{mAP on \% of labelled images}} \\
    \cline{2-7}
      ($\delta$)    & 7.5\% & 10\% & 12.5\% & 15\% & 17.5\% & 20\%\\
    \hline
    1  & 60.5$\pm$0.5 & 64.4$\pm$0.4 & 67.2$\pm$0.4 & 68.7$\pm$0.3 & 70.4$\pm$0.4 & 71.6$\pm$0.3  \\
    2  & 60.4$\pm$0.6& 65.8$\pm$0.4 & 67.6$\pm$0.3 & 69.0$\pm$0.4 & 70.6$\pm$0.2 & 71.5$\pm$0.3  \\
    3  & 60.5$\pm$0.6& 65.4$\pm$0.4 & 67.7$\pm$0.5 & 70.0$\pm$0.3 & 71.3$\pm$0.3 & 71.8$\pm$0.2  \\
    4  & 60.8$\pm$0.5& 66.2$\pm$0.4 & 68.4$\pm$0.2 & 70.5$\pm$0.4 & 71.6$\pm$0.3 & 72.6$\pm$0.1  \\
    5  & 61.2$\pm$0.5 & 65.0$\pm$0.6 & 68.0$\pm$0.4 & 70.1$\pm$0.5 & 71.4$\pm$0.2 & 72.6$\pm$0.1  \\
    6  & 59.8$\pm$0.7 & 65.9$\pm$0.5 & 67.9$\pm$0.3 & 70.1$\pm$0.3 & 71.4$\pm$0.2 & 72.3$\pm$0.2  \\
    \hline
    \end{tabular}}
  \end{center}
  \vspace{-0.6cm}
  \caption{\small Ablation study using VOC RetinaNet on how the budget expanding ratio $\delta$, which determines the size of the candidate pool in the first stage. The 1st round mAP is 43.4$\pm$2.2.}
\label{tbl:poolSize}
\vspace{-0.4cm}
\end{table}

\begin{figure*}[!ht]
    \centering
    \includegraphics[width=0.85\textwidth]{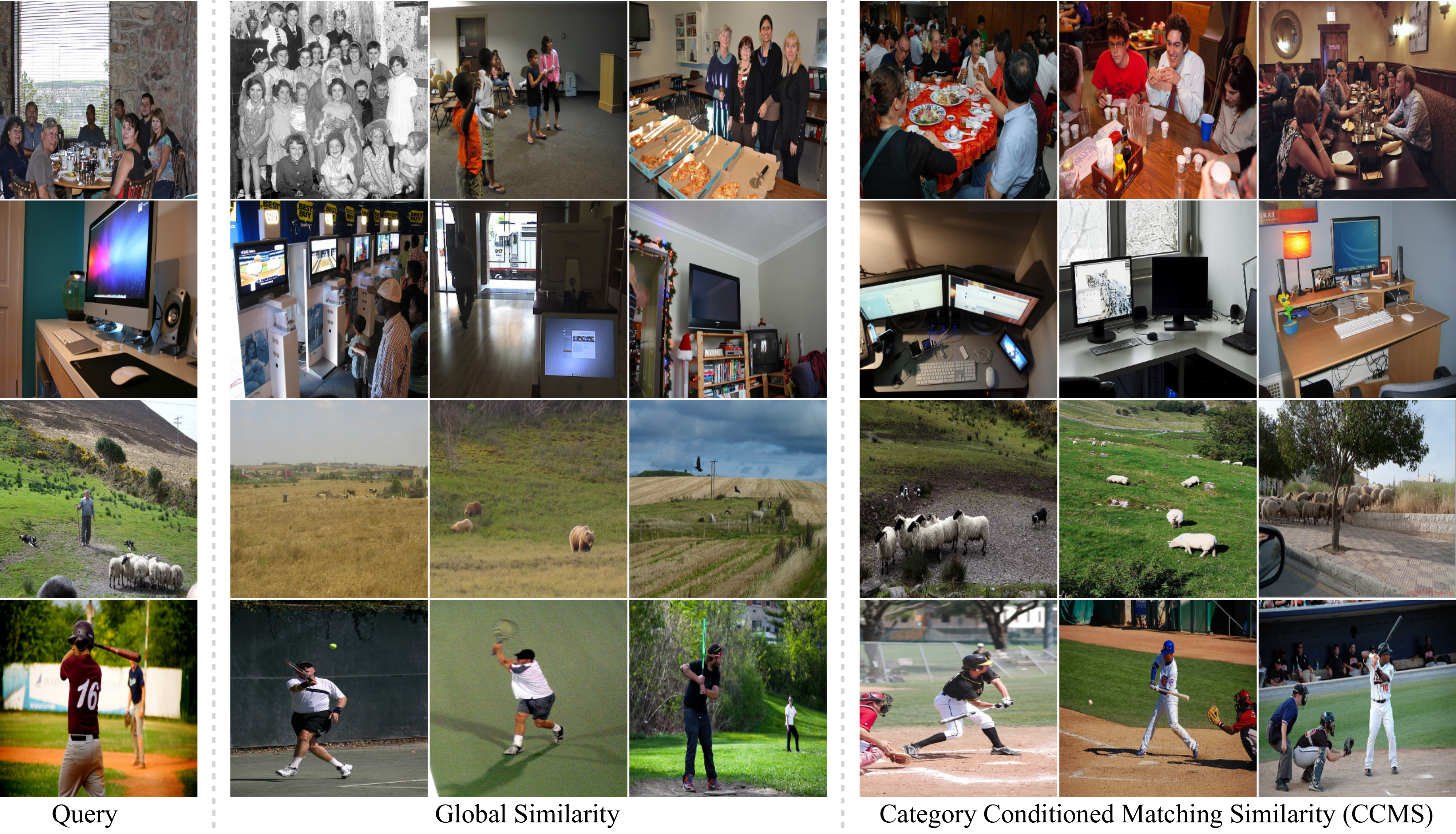}
    \vspace{-2mm}
    \caption{\small Image retrieval visualisation using global similarity and the proposed \ourSimShort~on COCO \textit{mini-val} using RetinaNet.}
    \label{fig:retrival}
    \vspace{-0.5cm}
\end{figure*}

\noindent\textbf{Size of the candidate pool.}
In Tab.\ref{tbl:poolSize}, we show how the budget expanding ratio $\delta$, which determines the size of the candidate pool, affects \ourmethodShort's performance using Pascal VOC RetinaNet. We observe that when $\delta$ is around 4 our method can achieve the best performance. Specifically, we find that a too large $\delta$, like $\delta \geq 6$, will harm \ourmethodShort's performance in early rounds, because in such cases the candidate pool will include many samples that the model is certain on, harming the overall information gain. On the other hand, a too-small expanding ratio, like $\delta < 3$, will harm \ourmethodShort's performance in later rounds because of the lacking of sample diversity in such small candidate pools.

\begin{table}[t]
  \begin{center}
  \resizebox{\linewidth}{!}{
    \begin{tabular}{c|cccccc}
    \hline
    \multirow{2}{*}{Uncertainty} & \multicolumn{6}{c}{mAP on \% of labelled images} \\
    \cline{2-7}
      & 7.5\% & 10\% & 12.5\% & 15\% & 17.5\% & 20\%\\
    \hline
    Random & 51.5$\pm$1.5 & 57.6$\pm$1.6 & 60.7$\pm$1.3 & 63.9$\pm$1.0 & 66.3$\pm$1.1 & 66.6$\pm$1.0 \\
    \hline
    Posterior  & 60.8$\pm$0.6 & 66.0$\pm$0.5 & 68.7$\pm$0.3 & 71.2$\pm$0.3 & 71.5$\pm$0.3 & 72.3$\pm$0.2  \\
    Margin  & 59.9$\pm$0.5 & 66.1$\pm$0.4 & 67.8$\pm$0.5 & 70.8$\pm$0.2 & 71.4$\pm$0.3 & 72.8$\pm$0.1  \\
    Entropy  & 60.8$\pm$0.5& 66.2$\pm$0.4 & 68.4$\pm$0.2 & 70.5$\pm$0.4 & 71.6$\pm$0.3 & 72.6$\pm$0.2  \\
    \hline
    \end{tabular}}
  \end{center}
  \vspace{-0.6cm}
  \caption{\small Comparison of different uncertainty measurements on VOC RetinaNet. The 1st round mAP is 43.4$\pm$2.2. }
  \vspace{-0.4cm}
\label{tbl:uncertainty}
\end{table}

\noindent\textbf{Uncertainty Measurement.}
As presented in Sec.\ref{sec:uncertainty}, in \ourmethodShort~we use entropy as the default uncertainty measurement. In Tab.\ref{tbl:uncertainty}, we show that our method can achieve similar performance when generalising to other types of uncertainty measurement. Specifically, we test \ourmethodShort~on Pascal VOC RetinaNet using two alternative uncertainty measurements: posterior probability~\cite{lewis1994heterogeneous} and probability margin~\cite{citovsky2021batch}. We observe that the posterior probability has a very close performance to entropy. However, the probability margin achieves an inferior performance in the early rounds, but can get similar performances with the other two uncertainty measurements in later rounds.

\noindent\textbf{Generalising to different backbones.}
In Tab.\ref{tbl:backbone}, we show our approach can well generalise to other backbone networks. We test \ourmethodShort~on Pascal VOC RetinaNet using the heavy-weighted ResNet-101~\cite{he2016deep}, light-weighted MobileNet v2~\cite{sandler2018mobilenetv2} and the high-performing transformer based SwinTransformer-Tiny~\cite{liu2021swin}. The results show that \ourmethodShort~is consistently better than random sampling by a large margin in all those backbone architectures, validating the strong generalisation ability of our approach. 

\begin{table}[t]
  \begin{center}
    \resizebox{\linewidth}{!}{
    \begin{tabular}{l|c|cccccc}
    \hline
    \multirow{2}{*}{Backbone} & \multirow{2}{*}{Method} &\multicolumn{6}{c}{\small{mAP on \% of labelled images}} \\
    \cline{3-8}
         & & 7.5\% & 10\% & 12.5\% & 15\% & 17.5\% & 20\%\\
    \hline
    \multirow{2}{*}{ResNet101} &Rand & 59.3$\pm$1.4 & 64.6$\pm$1.0 & 68.4$\pm$1.1 & 69.8$\pm$1.1 & 71.0$\pm$0.9 & 72.3$\pm$1.0  \\
                            &\ourmethodShort & 64.3$\pm$0.5& 69.6$\pm$0.3 & 71.1$\pm$0.2 & 72.8$\pm$0.3 & 73.8$\pm$0.1 & 75.0$\pm$0.1  \\ 
    \hline
    \multirow{2}{*}{MobileNetV2}  &Rand & 31.2$\pm$2.8 & 31.4$\pm$2.4 & 39.7$\pm$2.3 & 42.8$\pm$2.0 & 42.9$\pm$1.9 & 44.7$\pm$1.7  \\
                                 &\ourmethodShort& 35.7$\pm$1.7 & 41.9$\pm$1.2 & 45.6$\pm$1.3 & 46.3$\pm$1.2 & 48.4$\pm$0.9 & 50.7$\pm$0.9  \\ 
    \hline
    \multirow{2}{*}{SwinTiny}  &Rand &  63.8$\pm$1.0 & 67.9$\pm$0.9 & 70.0$\pm$1.0 & 71.2$\pm$0.6 & 71.8$\pm$0.8 & 73.3$\pm$0.6  \\
                               &\ourmethodShort  &  68.4$\pm$0.3 & 71.2$\pm$0.4 & 72.7$\pm$0.2 & 74.8$\pm$0.1 & 75.4$\pm$0.1 & 76.2$\pm$0.1  \\ 
    \hline
    \end{tabular}}
  \end{center}
  \vspace{-0.6cm}
  \caption{
  \small Comparison of \ourmethodShort~and random sampling on VOC RetinaNet using different backbones. The 1st round mAPs are 51.0$\pm$1.8, 20.9$\pm$3.4, and 59.0$\pm$1.5 for all three backbones.}
  \vspace{-0.4cm}
\label{tbl:backbone}
\end{table}

\begin{figure}[!t]
\centering
    \includegraphics[width=\linewidth]{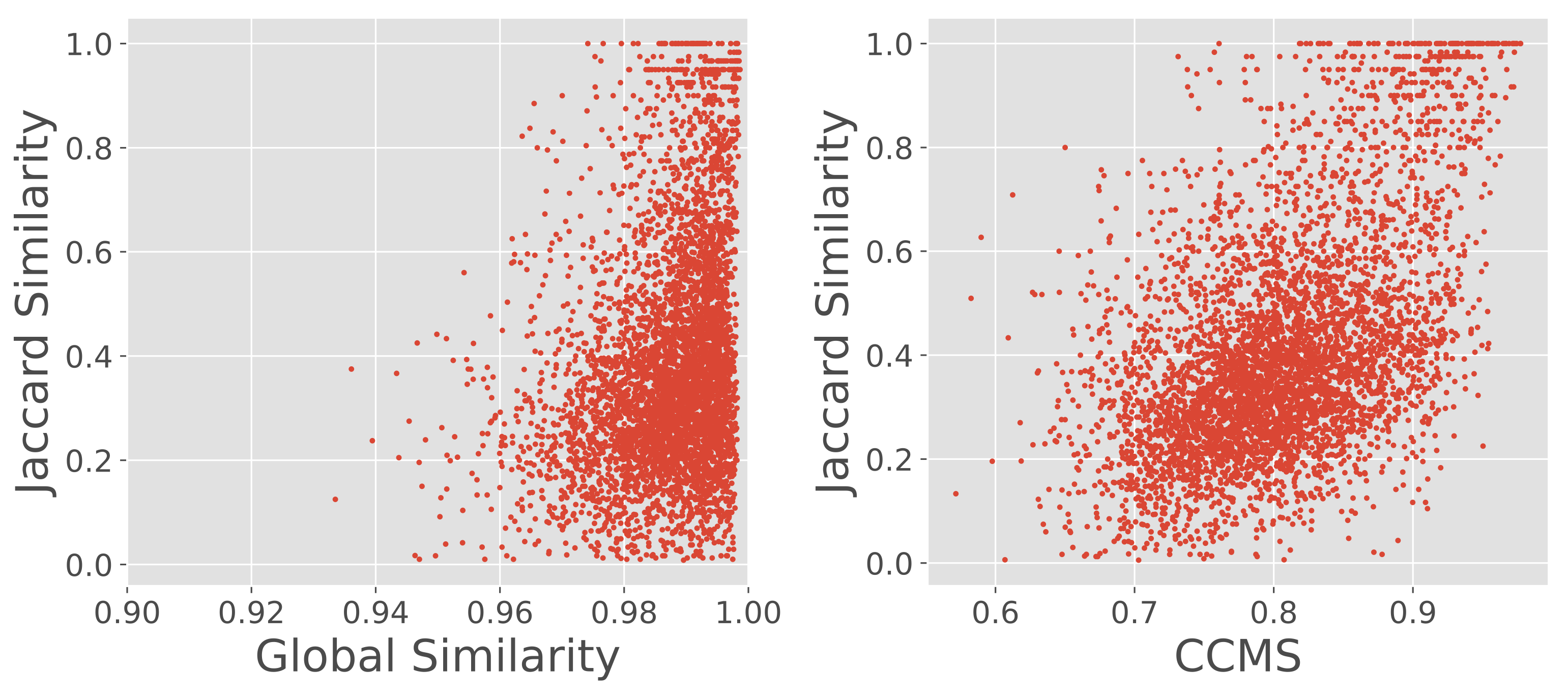}
    \vspace{-8mm}
    \caption{\small Comparison between global similarity and the proposed \ourSimShort~on image retrieval using RetinaNet on COCO \textit{mini-val}. We retrieve 20 most similar images for each anchor image. The similarity numbers are normalised to $(0,1)$.}
    \label{fig:jaccard}
    \vspace{-6mm}
\end{figure}

\noindent\textbf{Why is \ourSimShort~better than global similarity?}
Here we investigate why the proposed \ourSimShort~is better than global similarity in measuring image-wise similarity for multi-instance images by running an image retrieval experiment on COCO \textit{mini-val} set using a pre-trained RetinaNet detector: For each anchor image, we use both means of computing similarity to retrieve the 20 most similar images. Then we check whether the retrieved images depict similar scenes with the anchor image by measuring the averaged Jaccard similarity of object categories contained in those images. Specifically, for image $I_i$ and $I_j$ and their contained object category set $C_i$ and $C_j$, their Jaccard similarity is defined as $\mathcal{J}_{ij} = \frac{|C_i \cap C_j|}{|C_i \cup C_j|}$. For example, if an image contains $\{$dog, human$\}$ and the other image contains $\{$dog, cat$\}$, their Jaccard similarity $0.33$. We show the result on the whole COCO \textit{mini-val} set in Fig.\ref{fig:jaccard}. We observe that the averaged \ourSimShort~and Jaccard similarity are positively correlated, i.e., a high averaged \ourSimShort~usually corresponds to a high averaged Jaccard similarity and vice versa. However, global similarity does not hold such correspondence to the class-wise Jaccard similarity. In Fig.\ref{fig:retrival}, we show the image retrieval results by visualising the 3 most similar images with the anchor. From it we get two observations: (1) Global similarity is usually biased toward the dominating objects in the image while ignoring other objects (1st and 2nd rows) and \ourSimShort~is not; (2) \ourSimShort~is better than global similarity in capturing the fine-grained details in the image (3rd and 4th rows). These results validate our argument that \ourSimShort~is a more suitable similarity computing method for diversity-based active learning for object detection, which usually uses multi-instance images as input.

\vspace{-1mm}
\section{Conclusion}
We introduce a two-stage active learning algorithm for object detection. In the first stage, we propose \ourUncetaintySample~to select a candidate pool of uncertain samples,
and in the second stage we select a diverse query set using \ourSim. We show our method can generalise well and outperform previous works in various architectures and datasets.

\paragraph{Acknowledgements.} 
Chenhongyi Yang was supported by a PhD studentship provided by the School of Engineering, University of Edinburgh.

\clearpage

{\small
\bibliographystyle{ieee_fullname}
\bibliography{egbib}
}

\end{document}